\pdfoutput=1
\PassOptionsToPackage{dvipsnames,table,xcdraw}{xcolor}

\documentclass[11pt]{article}

\usepackage[]{acl}

\usepackage{times}
\usepackage{latexsym}

\usepackage[T1]{fontenc}

\usepackage[utf8]{inputenc}

\usepackage{microtype}

\usepackage{algorithm}

\usepackage[noend]{algpseudocode}
\usepackage{booktabs}
\usepackage{subcaption}
\usepackage{multirow}
\usepackage{url}
\usepackage{soul}
\usepackage{comment}
\usepackage{amsmath}
\usepackage{mathtools}
\usepackage{numprint}

\usepackage[textsize=scriptsize,backgroundcolor=green]{todonotes}

\newcommand{\tool}{{\sc {P}roto{TE}x}}
\newcommand{\toolsimple}{{\sc {S}imple{P}roto{TE}x}}
\title{\tool{}: {E}xplaining Model Decisions with {P}rototype {T}ensors}

\author{Anubrata Das$^1$\!\,\thanks{~\,Both authors contributed equally.}\ ~\ \
Chitrank Gupta$^{2\, *}$ \  
Venelin Kovatchev$^1$\ \ 
Matthew Lease$^1$ \ \
Junyi Jessy Li$^3$\\
$^1$School of Information ~and~ $^3$Dept.\ of Linguistics, The University of Texas at Austin\\
$^2$Dept.\ of Computer Science, Indian Institute of Technology Bombay\\
{\small \tt \{anubrata, venelin, ml, jessy\}@utexas.edu,}
{\small \tt chigupta2011@gmail.com}
}

\begin{document}
\maketitle
\begin{abstract}

We present \tool{}, a novel {\em white-box} NLP classification architecture based on prototype networks~\cite{li2018deep}. \tool{} faithfully explains model decisions based on prototype tensors that encode latent clusters of training examples. At inference time, classification decisions are based on the distances between the input text and the prototype tensors, explained via the training examples most similar to the most influential prototypes. We also describe a novel interleaved training algorithm that effectively handles classes characterized by the \emph{absence} of indicative features. On a propaganda detection task, \tool{} accuracy matches BART-large and exceeds BERT-large with the added benefit of providing faithful explanations. A user study also shows that prototype-based explanations help non-experts to better recognize  propaganda in online news.

\end{abstract}

\section{Introduction}
\label{sec:intro}

Neural models for NLP have yielded significant gains in predictive accuracy across a wide range of tasks. However, these state-of-the-art models are typically less interpretable than simpler, traditional models, such as decision trees or nearest-neighbor approaches. In general, less interpretable models can be more difficult for people to use, trust, and adopt in practice. Consequently, there is growing interest in going beyond simple ``black-box'' model accuracy to instead design models that are both highly accurate and human-interpretable.

\begin{figure*}[t!]
 \centering
 \includegraphics[width=0.9\textwidth]{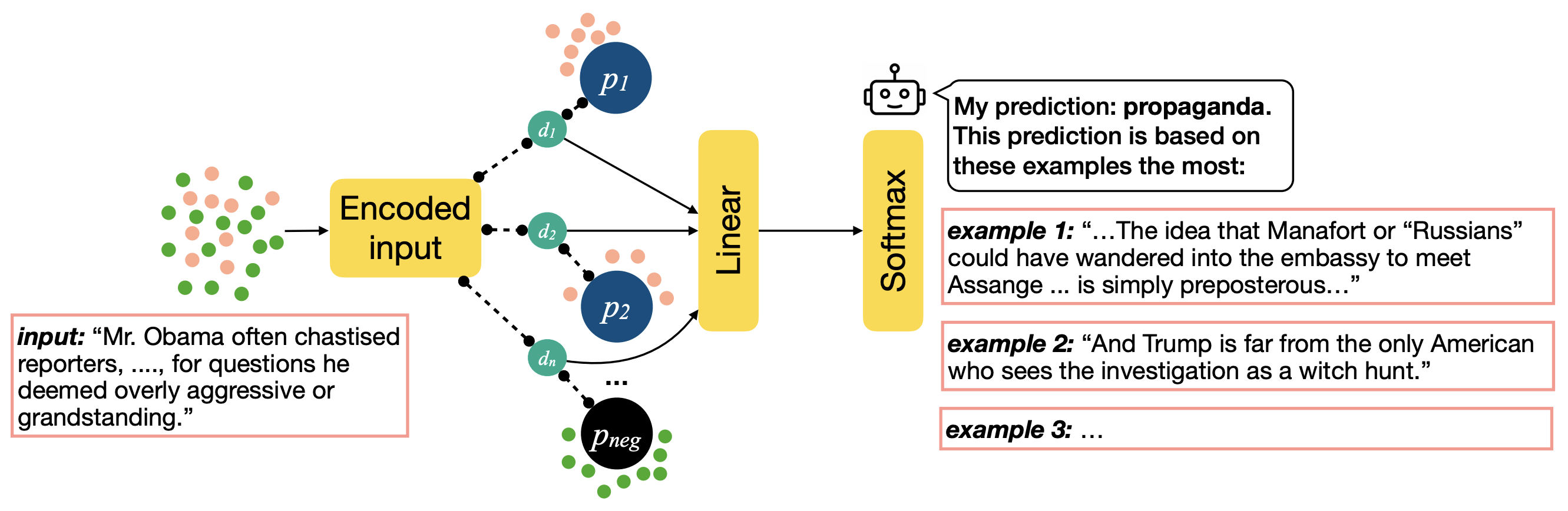}
 \captionof{figure}{\tool{} architecture along with a use case demonstration. Pink/Green dots denote training examples, which are clustered around positive prototypes (blue) and a single negative prototype (black). Dotted lines represent distances. In this use case, the user gives \tool{} an input, which produces a prediction while retrieving a set of highest ranked examples that \emph{directly} influenced model decision. In this diagram, by using ``overly aggressive or grandstanding'' the input creates propaganda via exaggeration~\cite{da2019fine}.
 \tool{} learns to identify sentences that contain propaganda phrases. In example 1, using {\em ``Russian''} to describe Manafort (Former American political consultant) constitutes propaganda, so does using {\em``witch hunt''} in example 2.
 Exposure to similar examples helps users build an intuition towards the language used in propaganda.} 
 \label{fig:architecture}
\vspace{-.5em}
\end{figure*}

While much research on white-box explainable models focuses on attributing parts of the input (e.g., word sequences) to a model's prediction~\cite{xu2015show,lei2016rationalizing,bastings2019interpretable,jain2020learning,glockner2020you}, there is much debate around their faithfulness and reliability~\cite{serrano2019attention,jain2019attention,wiegreffe2019attention,pruthi2020learning}. Additionally, while such local explanations (if faithful) can be extremely useful in more intuitive tasks such as sentiment classification, that may not be the case for difficult tasks where human judgments may require a high degree of training or domain expertise. In such cases, understanding how models make their decisions for a particular input based on its \emph{training data} can be insightful especially for engaging with users to develop an intuition on the model's decision making process.  



In this paper, we propose 
\textbf{Protot}ype {\bf T}ensor \textbf{Ex}plainability Network
(\tool{})\footnote{\url{https://github.com/anubrata/ProtoTEx/}} to faithfully explain  classification decisions in the tradition of case-based reasoning~\cite{kolodner1992introduction}. Our novel {\em white-box} NLP architecture augments prototype classification networks~\cite{li2018deep} 
with large-scale pretrained transformer language models.
Through a novel training regime, the network learns 
a set of prototype tensors that encode latent clusters of training examples.
At inference time, classification decisions are entirely based on  similarity to prototypes. 
This enables model predictions to be faithfully explained based on these  prototypes, \emph{directly} via similar training examples (i.e., those most similar to top-matched prototypes). We build upon the state-of-the-art NLP neural architectures to augment their accuracy with faithful and human-interpretable explanations. Figure~\ref{fig:architecture} shows an example of \tool{} on the task of \emph{propaganda detection} \cite{da2019fine}. 
%
%

Another contribution of \tool{} concerns effective modeling of positive vs.\ negative classes in the presence of asymmetry. In a typical binary classification (e.g., sentiment detection), the presence of positive vs.\ negative language can be used to distinguish classes. However, with a task such as Web search, what most distinguishes relevant vs.\ irrelevant search results is the presence vs.\ absence of relevant content. Having this absence (rather than presence) of certain features most clearly distinguish a class 
complicates both predicting it and explaining these predictions to users. 
To address this, we introduce a single {\em negative prototype} for representing the negative class, learned via a novel training regime. We show that including this negative prototype significantly improves results.  

While our model is largely agnostic to the prediction task, we evaluate \tool{} on a sentence-level binary propaganda detection task
\cite{da2019fine}. Recent work on explainable fact-checking \cite{kotonya2020explainable} has provided explanations via attention \cite{Popat2018DeClarEDF, shu2019defend}, rule discovery \cite{GadElrab2019ExFaKTAF}, and summarization \cite{Atanasova2020GeneratingFC, Kotonya2020ExplainableAF, kotonya2020explainable}, but not prototypes. Better explanations could enable support for human fact-checkers \cite{Nakov2021AutomatedFF}.

%

We show that \tool{} provides faithful explanations without reducing 
classification accuracy, which remains comparable to the underlying encoder, BART-large~\cite{lewis2020bart}, superior to that of BERT-large~\cite{devlin2019bert}, and with the added benefit of faithful explanations in the spirit of case-based reasoning. Furthermore, to the best of our knowledge, we are the first work in NLP that examines the utility of global case-based explanations for non-expert users in model understanding and downstream task accuracy. 



\section{Related work}
\vspace{-0.3em}

\paragraph{Explainable classification}
Unlike post-hoc analysis approaches for explainability~\cite{ribeiro2016should,sundararajan2017axiomatic}, prototype classification networks~\cite{li2018deep,chen2019looks,hase2019interpretable} are white-box models with explainability built-in via case-based reasoning~\cite{kolodner1992introduction} rather than extractive rationales~\cite{lei2016rationalizing,bastings2019interpretable,jain2020learning,glockner2020you}. They are the neural variant of prototype classifiers~\cite{bien2011prototype,kim2014bayesian}, predicting based on similar known instances.
Contemporary work~\cite{rajagopal-etal-2021-selfexplain} also stressed the importance of ``global'' explainability through training examples, yet in their approach, the similar training examples are not directly integrated in the decision itself; in contrast, we do so via learned prototypes to provide more transparency.
%

Our work builds on \citet{li2018deep}, which we lay out in Section~\ref{sec:prototype}. Later work \cite{chen2019looks,hase2019interpretable} enables prototype learning of partial images. In NLP,
\citet{guu2018generating} retrieved prototype examples from the training data for edit-based natural language generation.
\citet{hase2020evaluating} used a variant of \citet{chen2019looks}'s work to examine among other approaches; unlike our work, they used feature activation to obtain explanations similar to post-hoc approaches, and did not handle the absence of relevant content.



\paragraph{Evaluating explainability}

Explainability is a multi-faceted problem. HCI concerns include: 
a) For whom are we designing the explanations? b) What goals are they trying to achieve? c) How can we best convey information without imposing excessive cognitive load? and d) Can explainable systems foster more effective human+AI partnerships \cite{amershi2019guidelines, wickramasinghe2020trustworthy, wang2019designing, liao2020questioning, wang2021putting, bansal2021does}? On the other hand, algorithmic concerns include generating faithful and trustworthy explanations \cite{ jacovi2020towards}, local vs.\ global explanations, and post-hoc vs.\ self-explanations \cite{danilevsky2020survey}.

Explainability evaluation methods~\cite{doshi2017towards} include measuring faithfulness \cite{jacovi2020towards}, enabling model simulatability \cite{hase2019interpretable}, behavioral testing \cite{ribeiro2020beyond}, and evaluating intelligent user interactions \cite{nguyen2018believe}. 

\paragraph{Human+AI fake news detection}

While explainable fact-checking  \cite{kotonya2020explainable} could better support human-in-the-loop fact-checking \cite{Nakov2021AutomatedFF,demartini2020human}, studies rarely assess a human+AI team in combination 
\cite{nguyen2018believe}. In fact, human+AI teams often under-perform the human or AI working alone  \cite{bansal2021does}, emphasizing the need to carefully baseline  performance.


Propaganda detection \cite{da2019fine} constitutes a form of disinformation detection. 
Because propaganda detection is a hard task for non-expert users and state-of-the-art models are not accurate enough for practical use, explainability may promote adoption of computational propaganda detection systems \cite{martino2020survey}. 


\section{Methodology}

We adopt prototype classification networks \cite{li2018deep} first proposed for vision tasks as the foundation for our prototype modeling work (Section~\ref{sec:prototype}). 
We design a novel interleaved training procedure, as well as a new batching process, to (a) incorporate large-scale pretrained language models, and (b) address within classification tasks where some classes can only be predicted by the \emph{absence} of characteristics indicative of other classes.

\subsection{Base architecture}
\label{sec:prototype}

\tool{} is based on \citet{li2018deep}'s Prototype Classification Network, and we integrate pretrained language model encoders under this framework.
%
Their architecture is based on learning \emph{prototype tensors} that serve to represent latent clusters of similar training examples (as identified by the model). 
Classification is performed via a linear model that takes as an input the distances to the prototype tensors. 
As such, the network is a {\em white-box} model where global explanation is attained by \emph{directly} linking the model to learned clusters of the training data.

Shown in Figure~\ref{fig:architecture}, the input is first encoded into a latent representation.
This representation is fed through a prototype layer, where each unit of that layer is a learned prototype tensor that represents a cluster of training examples through loss terms $\mathcal{L}_{p1}$ and $\mathcal{L}_{p2}$ (specified by equations~\ref{eq:lp1} and~\ref{eq:lp2} below). 

For each prototype $j$, the prototype layer calculates the L2 distance between its representation $\mathbf{p_j}$ and that of the input $\mathbf{x}_i$, i.e., $||\mathbf{x}_i-\mathbf{p}_j||^2_2$. The output of the prototype layer, which is a matrix of L2 distances, is then fed into a linear layer; this learns a weight matrix of dimension $K\times m$ for $K$ classes and $m$ prototypes, where the $K$ weights learned for each prototype indicates that prototype's relative affinity to each of the $K$ classes. Classification is performed via softmax.




The total loss is a weighted sum of three terms:\footnote{In \citet{li2018deep}, a fourth reconstruction loss is used with their convolutional network. We found that incorporating a reconstruction loss  led to unstable training, so we omit it.}
\vspace{-0.5em}
\begin{equation}
\mathcal{L} = \mathcal{L}_{ce}
+\lambda_1 \mathcal{L}_{p1}
+\lambda_2 \mathcal{L}_{p2}
\label{eq:loss}
\vspace{-0.5em}
\end{equation}
with hyperparameter $\lambda$s, standard classification cross-entropy loss $\mathcal{L}_{ce}$, and two prototype loss terms, $\mathcal{L}_{p1}$ and $\mathcal{L}_{p2}$.
%

$\mathcal{L}_{p1}$ minimizes  avg.\ squared distance between each of the $m$ prototypes and $\ge 1$ encoded input:
\vspace{-0.5em}
\begin{equation}
\mathcal{L}_{p1} = \frac{1}{m}\sum_{j=1}^m      \min_{{i=1, n}} ||\mathbf{p}_j - \mathbf{x}_i||^2_2
\label{eq:lp1}
\end{equation}
encouraging each learned prototype representation to be similar to at least one training example.

$\mathcal{L}_{p2}$ encourages training examples to cluster around prototypes in the latent space by minimizing the average squared distance between every encoded input and at least one prototype:
\vspace{-0.5em}
\begin{equation}
\mathcal{L}_{p2} = \frac{1}{n}\sum_{i=1}^n     \min_{{j=1, m}} ||\mathbf{x}_i - \mathbf{p}_j||^2_2
\label{eq:lp2}
\end{equation}

\citet{li2018deep} used convolutional autoencoders to represent input images. 
However, in the context of NLP, convolutional neural networks do not have sufficient representation power ~\cite{elbayad2018pervasive} and transformer-based language models, which are pretrained on large amounts of data, have consistently performed better in recent research. 
Thus to encode inputs, we experiment with two such models: 
BERT~\citep{devlin2019bert} (a masked language model) and BART~\citep{lewis2020bart} (a sequence-to-sequence autoencoder). 


\paragraph{Intuition \& explainability based on case-based reasoning.} Because learned prototypes occupy the same space as encoded inputs, we can directly measure the distance between prototypes and encoded train or test instances. During inference time, prototypes closer to the encoded test example become more ``activated'', with larger weights from the prototype layer output. Consequently, model prediction is thus the weighted affinity of each prototype to the test example, where each prototype has $K$ weights over the possible class assignments.

In the context of classification in NLP, we operationalize case-based reasoning \cite{kolodner1992introduction} by providing similar training examples. Once the model is trained, for each prototype we rank the training examples by proximity in the latent space. During inference, we rank the prototypes by proximity to the test example. Thus, for a test example, we can obtain the training examples closest to the prototypes most influential to the classification decision. \citet{jacovi2020towards} define {\em faithfulness} as ``how accurately [explanations] reflects the true reasoning process of the model.'' Since prototypes are directly linked to the model predictions via a linear classification layer, explanations derived by the prototypes are faithful by design. We also provide a mathematical intuition of how prototype layers relates to soft-clustering (which is inherently interpretable) in the appendix \ref{subsec:math}.

\begin{algorithm}[t]
\small
\begin{algorithmic}[1]
\State$\mathbf{p}  \coloneqq \{p_{1}...p_{m}\}$
\Comment{prototypes}
\State$\mathbf{x} \gets \text{Encoder}(s_1,s_2,...s_n)$
\Comment{encode input sentences}
\State Init$(\mathbf{p})$ 
\State\emph{LinearLayer}$\gets$ XavierInit
\For{$k$ iterations} 
        \For{batch $\mathbf{x}_b$ \textbf{in} Train}
            \State \emph{d} $\gets$ distance$(\mathbf{x}_{b}, \mathbf{p})$
            \State $ \mathcal{L}_{ce} \gets$ CE$(\text{\it LinearLayer}(\text{norm}(d),\mathbf{y}_b)) $
            
            \State $ loss \gets \mathcal{L}_{ce}+ \lambda_1 \mathcal{L}_{p1}(d)+ \lambda_2 \mathcal{L}_{p2}(d)$
            
            \State Update($Encoder, LinearLayer$,\textbf{p} ; $loss$)
        \EndFor
\EndFor
\end{algorithmic}
\caption{\footnotesize{Training for \toolsimple}.}
\label{alg:trainingsimple}
\end{algorithm}

\begin{algorithm}[t]
\small
\begin{algorithmic}[1]
\State$\mathbf{p}_{\text{pos}}  \coloneqq \{p_{1}...p_{m-1}\}$
\Comment{prototypes for  $\oplus$ class}
\State$\mathbf{p}_{\text{neg}}$
\Comment{single prototype for $\ominus$ class}
\State$\mathbf{x} \gets \text{Encoder}(s_1,s_2,...s_n)$
\Comment{encode input sentences}

\State Init$(\mathbf{p}_{\text{pos}},\mathbf{p}_{\text{neg}})$ 
\State\emph{LinearLayer}$\gets$ XavierInit
\For{$k$ iterations} 
    \For{$i \in 1\!:\!\delta$ epochs} 
    \Comment{Minimize $\mathcal{L}_{p1}$ loss}
    \State $c \gets \ i\!\!\!\mod{}\!2$ \Comment{pick $\{\oplus,\ominus\}$ class this iteration}
    \State $\mathbf{p}_c \gets$ prototype(s) for selected class \emph{c}
        \For{batch $\mathbf{x}_b$ \textbf{in} Train}
            \State $\mathbf{d}_c \gets$ distance$(\mathbf{x}_{bc}, \mathbf{p}_{c})$, \  $\mathbf{x}_{bc} \mkern-4mu \subset \! \textnormal{class} \ c$
            
            \State Update$(\mathbf{p}_c;\mathcal{L}_{p1}(\text{norm}(\mathbf{d}_c)))$
        
       \EndFor
    \EndFor
    \For{$j \in 1\!:\!\gamma$ epochs} 
    \Comment{Minimize $\mathcal{L}_{ce}$ \& $\mathcal{L}_{p2}$ loss}

        \State $c \gets \ j\!\!\!\mod{}\!2$ \Comment{pick $\{\oplus,\ominus\}$ class this iteration}
        \State $\mathbf{p}_c \gets$ prototype(s) for selected class \emph{c}
        \For{batch $\mathbf{x}_b$ in Train}
            \State $\mathbf{d}_c \gets$ distance$(\mathbf{x}_{bc}, \mathbf{p}_{c})$, \ $\mathbf{x}_{bc} \mkern-4mu  \subset \!$\ class $c$
            \State $\mathbf{d} \gets$ distance$(\mathbf{x}_{b}, \mathbf{p}_{\text{pos}},\mathbf{p}_{\text{neg}})$
            \State $ \mathcal{L}_{ce} \gets$ CE$(\text{\it LinearLayer}(\text{norm}(\mathbf{d}),\mathbf{y}_b)) $
            
            \State $ loss \gets \mathcal{L}_{ce}+ \lambda \mathcal{L}_{p2}(\text{norm}(\mathbf{d}_c))$
            
            \State Update(\emph{Encoder, LinearLayer; loss})
        \EndFor
    \EndFor
\EndFor
\end{algorithmic}
\caption{\footnotesize{Decoupled} training for prototypes and classification, which enables the learning of the negative prototype.}
\label{alg:training}
\end{algorithm}

\subsection{Handling asymmetry: {\em negative prototype}} \label{sec:negproto}

Section \ref{sec:intro} noted a challenge in effectively modeling positive vs.\ negative classes in the presence of asymmetry. With detection tasks (e.g., finding relevant documents \cite{Kutlu20-jair} or propaganda \cite{da2019fine}), the negative class may be most distinguished by the \emph{lack} of positive features (rather than presence of negative ones). If a document is relevant only if it contains relevant content, how can one show the lack of such content? 
This poses a challenge both in classifying negative instances and in explaining such classification decisions on the basis of missing features. 

For propaganda, \citet{da2019fine} side-step the issue by only providing rationales for positive instances. For relevance, \citet{Kutlu20-jair} define a {\em negative rationale} as summarizing the instance, to succinctly show it is not germane to the positive class. However, if we conceptualize the positive class as a specific {\em foreground} to be distinguished from a more general {\em background}, such ``summary`` negative rationales drawn from the background distribution are likely to provide only weak, noisy evidence for the negative class.

We investigate the potential value of including or excluding a single {\em negative prototype} to model this ``background'' negative class, and design an interleaved training procedure to learn this prototype.


\subsection{Training}
\label{susbsec:training}


We present two algorithms for training. The vanilla one, which we call {\toolsimple}, does not interleave the training of positive and negative prototypes. This is illustrated in \textbf{Algorithm \ref{alg:trainingsimple}}.

One of our contributions is the design of an iterative, interleaved approach to training that balances competing loss terms, encouraging each learned prototype to be similar to at least one training example ($\mathcal{L}_{p1}$) and encouraging training examples to cluster around prototypes ($\mathcal{L}_{p2}$).
We perform each type of representation update separately to ensure that we progressively push the prototypes and the encoded training examples closer to one another.

We illustrate this process in {\bf Algorithm~\ref{alg:training}}. We initialize prototypes with Xavier, which allows the prototype tensors to start blind (thus unbiased) with respect to the training data and discover novel patterns or clusters on their own. After initialization, in each iteration, we first update the prototype tensors to become closer to at least one training example (henceforth $\delta$ loop). Then, in a separate training iteration, we update the representations of the training examples to push them closer to the nearest prototype tensor (henceforth $\gamma$ loop). Since prototypes themselves do not have directly trainable parameters, we train the classification layer together with the encoder representations during the $\gamma$ loop. 
%
We further separate the training of the positive and negative prototypes in order to push the negative ``background'' examples to form its own cluster. To this end, we perform class-level masking by setting the distances between the examples and prototypes of different classes to \texttt{inf}.

Finally, we perform instance normalization~\cite{ulyanov2016instance} for all  distances in order to achieve segregation among different prototypes (namely, the prototypes of the same class do not rely solely on a handful of examples). We discuss the effects of instance normalization in Section~\ref{sec:results:classification}.


\begin{figure}
    \centering
    \includegraphics[width=0.9\columnwidth]{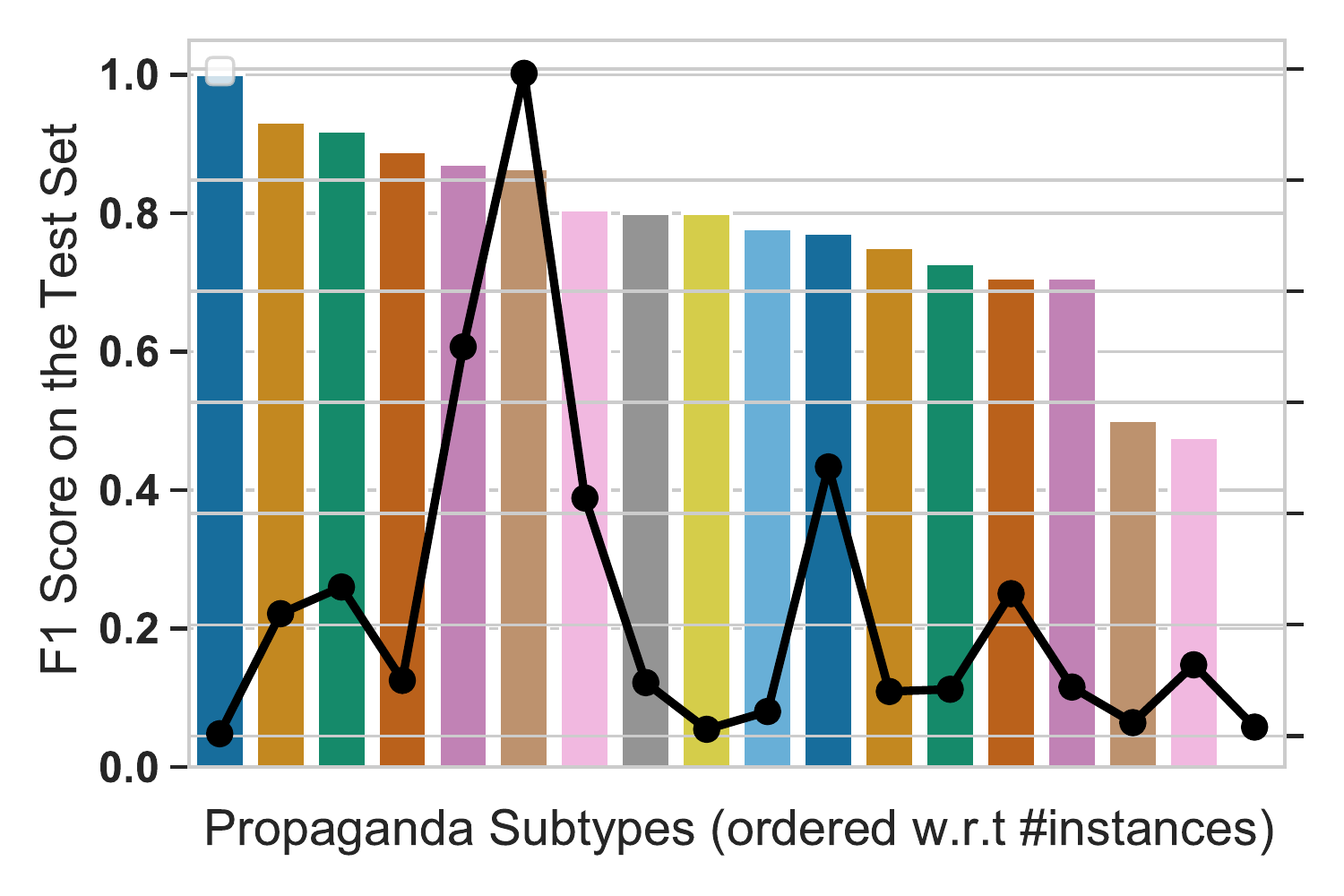}
    \caption{Macro-F1 score of \tool{} predicting examples that belong to each propaganda subclass. The black line corresponds to the number of examples in the test set. Classes are reordered in terms of F1.}
    \label{fig:subclass_f1}
\end{figure}

\section{Experiments}

\paragraph{Task}
We evaluate a binary sentence-level classification task predicting whether or not each sentence contains propaganda. We adopt \citet{da2019fine}'s dataset of 21,230 sentences from news articles, with a 70/10/20 train/development/test split. Only 35.2\% of sentences contain propaganda. The data is further classified into 18 fine-grained categories of propaganda; see analysis of prototypes in Section~\ref{sec:results:classification}.


\subsection{Models and Settings}
\label{subsec:model}
Hyperparameters are tuned on the validation data. Optimization for all neural models use AdamW~\cite{loshchilov2018decoupled} algorithm with a learning rate of 3e-5 and a batch size of 20.
We use early-stopping ~\cite{pytorch-ignite} with Macro F1 on validation data.
We further perform upsampling within each batch to balance the number of examples in the positive and the negative classes.

\begin{table}[t!]
\centering
\small
\npdecimalsign{.}
\nprounddigits{2}
\begin{tabular}{p{0.4\linewidth}|n{1}{3}n{1}{3}n{1}{3}}
\toprule
                            &  \text{Neg} $\ominus$    & \text{Pos} $\oplus$     & \text{Macro} \\
\midrule
Random                      &  0.5998393833    & 0.3429558633    & 0.4713976233 \\
\arrayrulecolor{black!60}\midrule
BERT-large                        & 0.8643608    & 0.58634953     & 0.725355165   \\
BART-large                  & 0.85840085    & 0.65    & \textbf{0.75}  \\
\arrayrulecolor{black!60}\midrule
KNN-BART(-large)                         & 0.82          & 0.52          & 0.67          \\
\toolsimple{}             & 0.83271852    & 0.39745812     & 0.61508832  \\
\tool{} (-norm) & 0.81514556    & 0.56273764    & 0.6889416   \\
\tool{} (+norm)                  & 0.85365422    & 0.64    & \textbf{0.75}    \\
\bottomrule
\end{tabular}
\caption{F1 measures on 
the task of propaganda detection. \tool{} performs similar to BART.}
\label{tab:results}
\vspace{-1em}
\end{table}

\paragraph{Prototype Models}
\tool{} can be used across different 
underlying encoders on which interpretability components are added. Empirically, we found BART performed better on classification and so adopt it. 
We empirically determine the optimal number of prototypes to be 20, with one negative prototype. $\delta=1,\lambda=2,\gamma_1=\gamma_2=0.9$. To achieve the maximum transparency, we set the bias term in the linear layer to 0 so that all information goes through the prototypes.\footnote{
~Early experiments showed no difference between including vs.\ excluding the bias term with instance normalization.} Additionally, we compare to \toolsimple{}, which trains without use of the negative prototype. 


\paragraph{Baselines}
As a strong \emph{blackbox} benchmark we use pretrained LMs without prototypes. 

{\bf BERT-large}~\cite{devlin2019bert}: 
we use a simple linear layer over the output of the \texttt{CLS} token from the BERT encoder for classification. 

{\bf BART-large}~\cite{lewis2020bart}: we use the \texttt{eos} token's representation from the BART encoder as input to the linear layer of the  model. 

We also include a random baseline
and a case-based reasoning K-Nearest-Neighbor (\textbf{KNN-BART}) baseline with the BART-large encoder.

\begin{figure}[t]
\centering

  \includegraphics[width=0.45\textwidth]{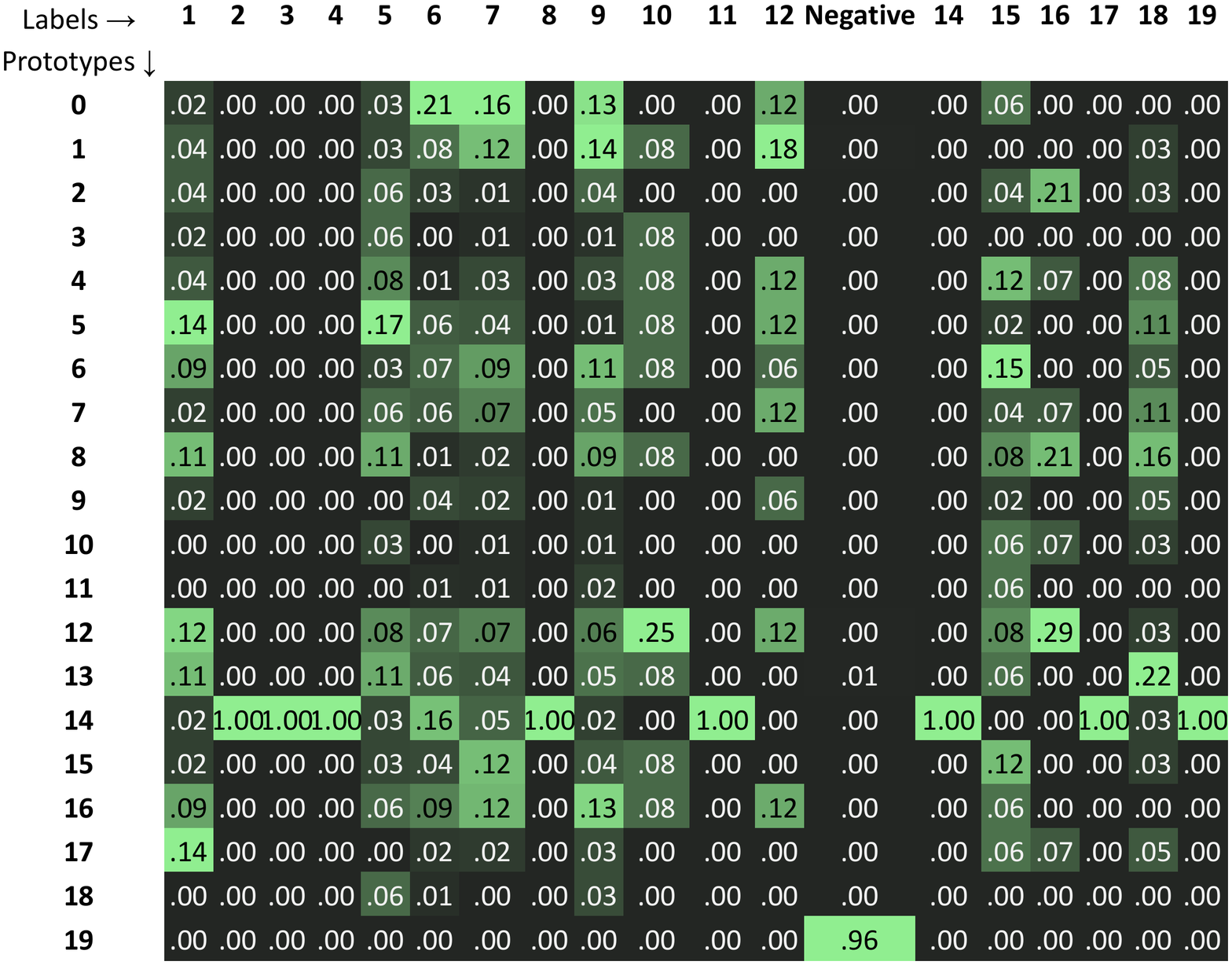}%


\caption{For each subcategory of propaganda (and the $\ominus$ class), the fraction of validation examples from that subcategory that are associated with each prototype; ``association'' defined as the closest prototype for that example. We see that \tool{} learns prototypes that ``focusses'' differently on the subcategories.
}
\label{fig:ablation:biasdrop}
\end{figure}

\subsection{Classification Results}\label{sec:results:classification}
{\bf Table~\ref{tab:results}} shows F1 scores achieved by models. 
Among the black-box baselines, the BART-large encoder representation outperformed BERT-large significantly ($p<0.05$, bootstrap test~\cite{berg2012empirical}).
\tool{} performed on-par with its underlying encoder BART, showing that \tool{}'s explainability came at no cost of classification performance. It also substantially outperforms the KNN-BART baseline.

{\bf Figure~\ref{fig:subclass_f1}}  shows F1 for the examples, pretaining to each subclass labeled by \citet{da2019fine}. We can see that the model performance is relatively consistent across subclasses. The two subclasses that are most difficult for the model are ``Reductio ad Hitleru'' and ``Appeal to Authority''.

In {\bf Figure \ref{fig:ablation:biasdrop}}, we visualize and show that
different prototypes ``focus'' on each subclass differently. We also see that negative examples are associated only with the negative prototype, and vice-versa.

\noindent\textbf{Negative Prototype.} Using a negative prototype slightly improves \toolsimple{} results. Lacking a negative prototype, the only way to classify a negative class would be via a negative correlation on the distance between the test input and the learned prototypes. The use of the negative prototype simplifies the discriminatory process by dissociating the classification process of the negative class from the classification process of the positive class.

\noindent\textbf{Instance Normalization.}
As shown in Table~\ref{tab:results}, normalization boosts classification performance. 
We also observe its benefit for explainability.

Because \tool{}'s explainability comes from retrieving the most informative training examples, it will not be helpful for people if all prototypes are close to only a few training examples. Instead, it would be more beneficial for the prototypes to represent more subtle patterns within the training examples belonging to the same class. We refer to this phenomenon as prototype \emph{segregation}.
While the classification layer ensures that positive and negative examples (and their prototypes) are separated, it does not take into account segregation \emph{within} the positive class. Similarly, the prototype losses $\mathcal{L}_{p1}$ and $\mathcal{L}_{p2}$ only locally ensure the closeness of examples to prototypes and vice-versa. To encourage segregation, we perform instance normalization~\cite{ulyanov2016instance} for all distances.

\begin{figure}[t]
    \centering
    \includegraphics[width=0.8\linewidth]{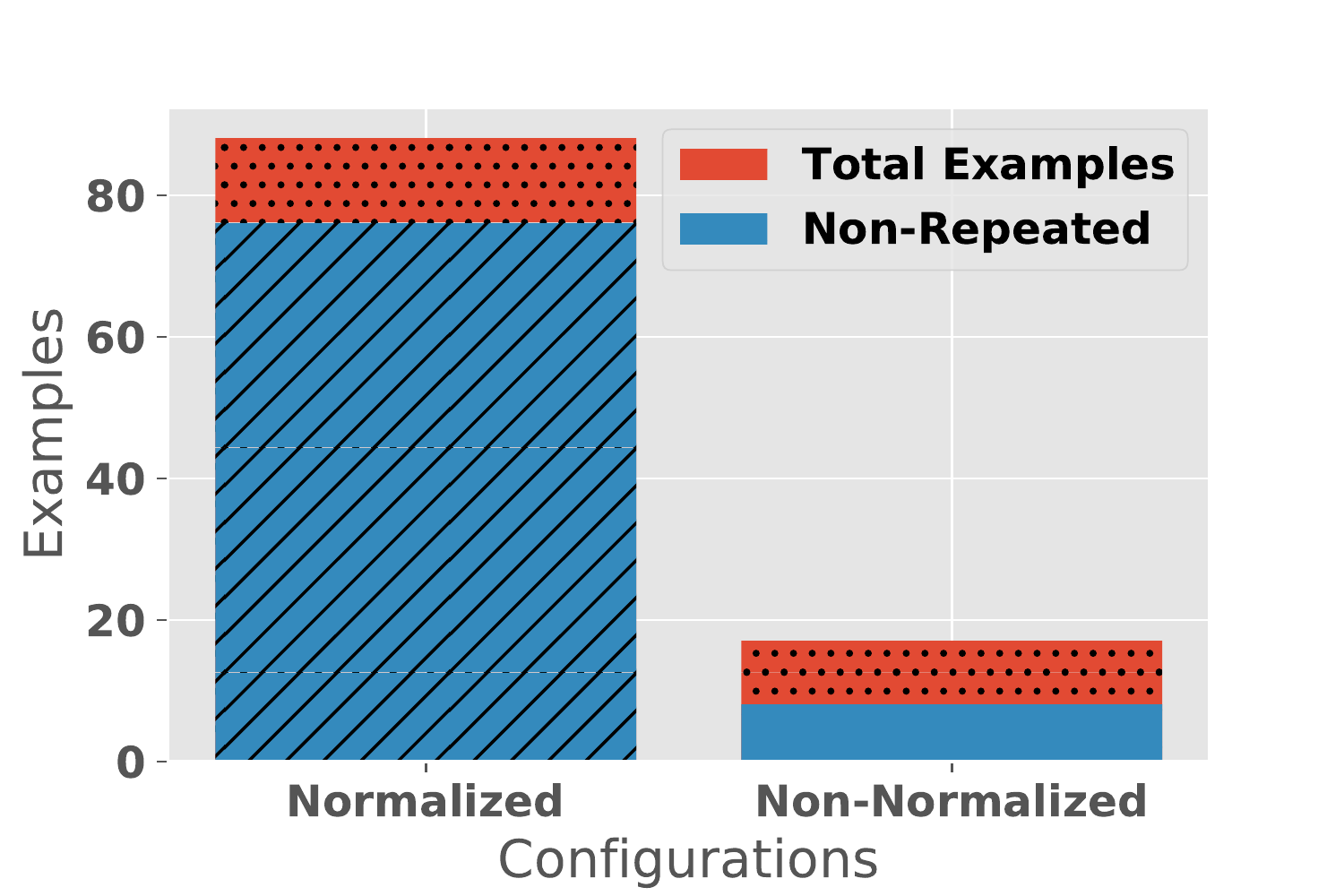}
    \caption{Number of unique 5-nearest training examples to each prototype (blue+red), and the number of examples associated with only 1 prototype (blue-only). Without normalization, very few examples (out of 100) are close to \emph{all} prototypes; with normalization, we observe more diversity: different training examples are near different prototypes.}
    \label{fig:norm_effect}
\end{figure}

\begin{table*}[t!]
\centering
\resizebox{\textwidth}{!}{%
\begin{tabular}{@{}llll@{}}
\toprule
\begin{tabular}[c]{@{}l@{}}\textbf{Input} \\ \textbf{Sentence}\end{tabular} &
  \begin{tabular}[c]{@{}l@{}}\textbf{True}\\ \textbf{Label}\end{tabular} &
  \begin{tabular}[c]{@{}l@{}}\textbf{Model}\\ \textbf{Prediction}\end{tabular} &
  \begin{tabular}[c]{@{}l@{}}\textbf{Similar}\\ \textbf{Examples}\end{tabular} \\ \midrule
\begin{tabular}[c]{@{}l@{}}"This  scandal  has  set  off  a  \hl{feeding  frenzy} as  \\ Internet  sleuths  search  for  other  incidents \\ in  which  Franken  has  acted  inappropriately."\end{tabular} &
  Propaganda &
  Propaganda &
  \begin{tabular}[c]{@{}l@{}}- "And  Trump  is  far  from  the  only  American  \\ who  sees  the  investigation  as  \hl{a  witch  hunt}. (\textbf{Name Calling,Labeling})"\end{tabular} \\
 &
   &
   &
  \begin{tabular}[c]{@{}l@{}}- "Do  the  FBI  and  law  enforcement  think  people \\ won’t  talk  about  it  or  speculate  as  to  what  happened? (\textbf{Doubt})"\end{tabular} \\
 &
   &
   &
  \begin{tabular}[c]{@{}l@{}}- "And  \hl{the  father  of  Muslim  spy}  ring  Imran  Awan  \\ transferred  a  USB  drive  to  a  Pakistani  senator  and  \\ former head of a  Pakistani intelligence agency." (\textbf{Name Calling,Labeling})\end{tabular} \\ \bottomrule
\end{tabular}%
}
\caption{Examples of similar sentences identified by our model. The input sentence uses the phrase {\em feeding frenzy} which is an example of propaganda phrasing. The model identifies training examples that also contain propaganda phrases as \hl{highlighted}. Note that the model does not obtain the highlights shown here. Highlights are also not part of our human evaluation.} 
\label{tab:exp-proto}
\end{table*}

This effect is shown in Figure~\ref{fig:norm_effect}. Specifically, we retrieve the 5 closest training examples for each of our 20 prototypes; good segregation would mean that a large portion of these examples are unique examples (the highest value is 100 meaning that all examples are unique), while bad segregation means that a large portion of these examples are the same (the lowest value is 5 meaning that all prototypes are the closest to only 5 training examples). Without normalization, we have only 17 unique examples for all 20 prototypes, yet with normalization this number is 88. Furthermore, almost all of the 88 training examples are associated with only one prototype.

\section{Human Evaluation}

\tool{} is designed to provide faithful case-based explanations (as shown in {\bf Table \ref{tab:exp-proto}}) for its classification decisions. Given the set of top prototypes most influential in predicting the class for a given example, we hypothesize that these top prototypes will be representative of the example and the label corresponding to the example. 
We carry out two user studies to assess the utility of these prototype-based explanations for non-expert end users. Specifically, we examine
whether model explanations help non-expert users to: 1) better recognize propaganda in online news; and 2) better understand model behavior. 





We obtain 540 user-responses, based on 20 test-set examples, 
balancing gold labels and model predictions to include 5 examples from each group: true-positives, false-negatives, true-negatives, and false-positives. To simplify propaganda definitions for non-experts, we pick only four types of propaganda and we provide participants with definitions and examples for each type: {\em Appeal to Authority}, {\em Exaggeration or Minimisation}, {\em Loaded Language}, and {\em Doubt}. 
We select these categories because they cover the majority of the examples in the test set. 

For each example, we select the top-5 prototypes that most influenced the model's prediction. We then represent each prototype by the closest training example in the embedding space. 
As with case-base reasoning, we explain model decisions to participants by showing for each test example the five training examples that best represent the evidence (prototypes) consulted by the model in making its prediction. Participants are primed that the model is wrong in 50\% of the cases (to prevent over-trust).



\begin{figure}[t!]
    \includegraphics[width=0.45\textwidth]{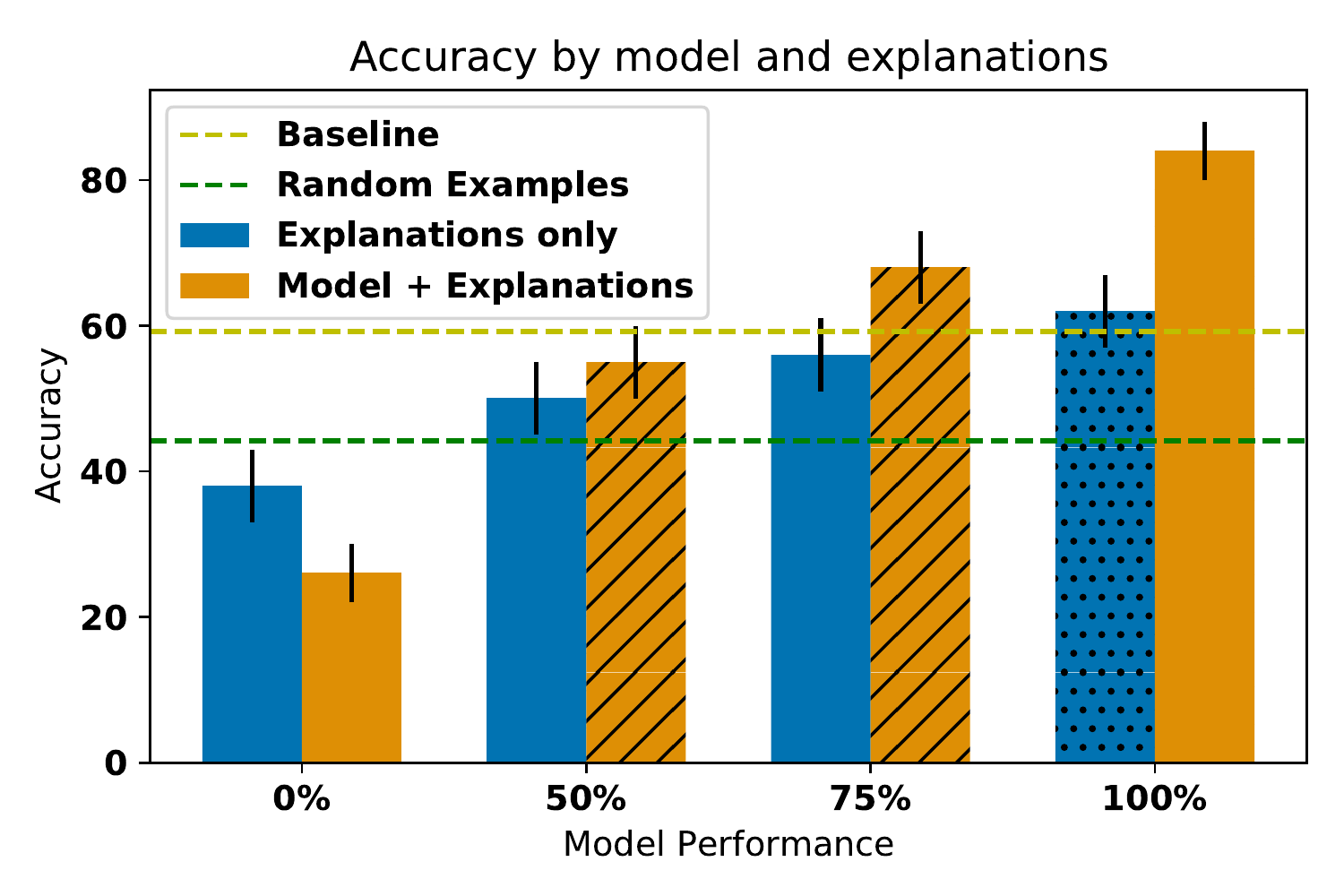}
    \caption{Accuracy of human annotations when provided with \tool{} explanations or \tool{} explanations + prediction. \textbf{Model Performance}: the accuracy of the model generating the explanations. \textbf{Baseline}: Annotation accuracy without explanations. \textbf{Random}: Randomly selected examples for explanation.
    }
    \label{fig:protovisual}
\end{figure}

\begin{figure*}[h!]
\centering
\subfloat[]{%
  \includegraphics[width=0.4\textwidth]{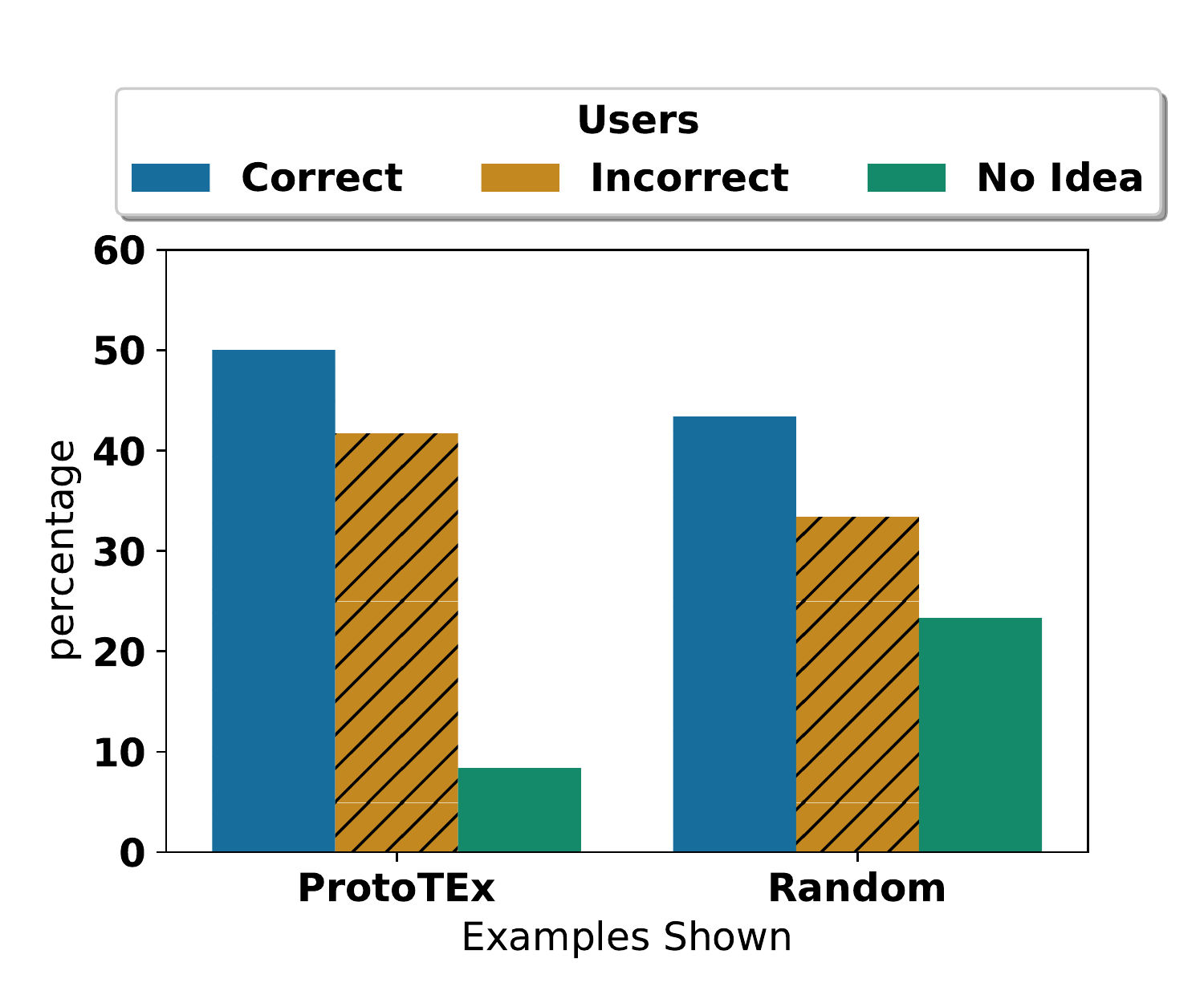}%
  \label{fig:mu}%

}
\subfloat[]{%
  \includegraphics[width=0.4\textwidth]{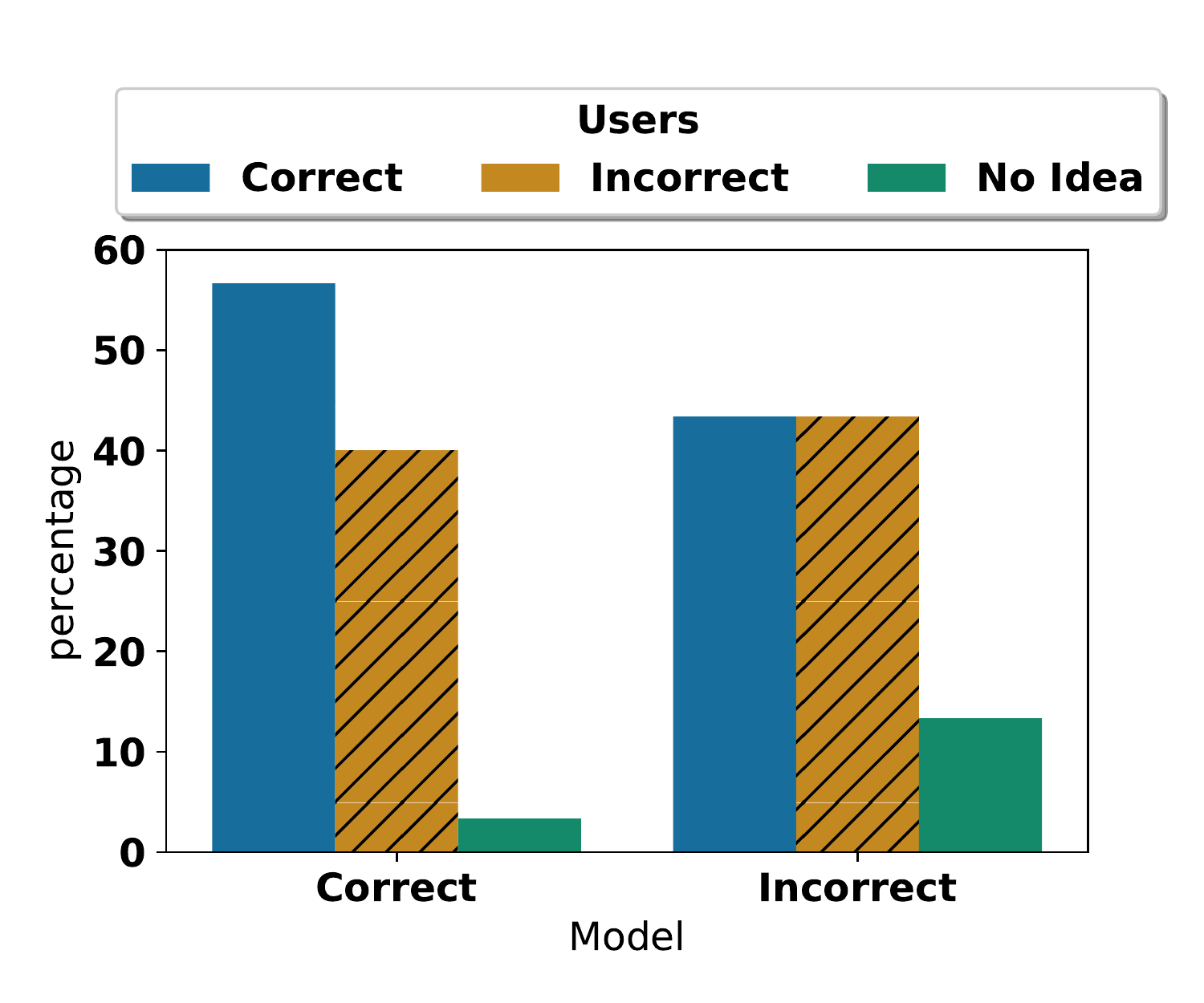}%
  \label{fig:mu-breakdown}%
}

\caption{Model Simulatability. User assessment of the model prediction \textbf{a)} Comparing \tool{} selected training examples vs.\ random examples; \textbf{b)} Comparing examples where the model prediction is accurate and examples where the model prediction is wrong}
\label{fig:model-understanding}
\vspace{-.5em}

\end{figure*}

\subsection{Recognizing Propaganda}
\label{recogProp}
In this first likert-scale rating task, participants are asked whether the test example contains propaganda. Options included: definitely, probably, probably not, definitely not, or ``I have no idea (completely unsure how to respond)''.
%
We compare the following four study conditions:

\textbf{No Explanation (Baseline)} We show only the test example that needs to be classified.

\textbf{Random Examples} We show five randomly selected training examples\footnote{Random sampling of examples has been successfully used in tasks such as Semantic Textual Similarity (STS) and Natural Language Inference (NLI) \cite{agirre-etal-2013-sem,gold-etal-2019-annotating} to obtain a reasonable lower-bound. Comparison with random baseline demonstrates that our system selects examples that can improve human performance.}. 

\textbf{Explanation Only (EO)} We also show five training examples, each representing a top-5 prototype influencing the model prediction, as the evidence consulted by the model in arriving at its prediction.

\textbf{Model Prediction + Explanations (ME)} Both the model prediction and explanations are shown.

\paragraph{Results} As Figure \ref{fig:protovisual} shows, 
in the first baseline condition (without any additional information), participants were able to correctly predict the presence of propaganda in 59\% of the cases. In the second baseline condition, when we providing random examples as ``explanation'', accuracy drops to 44\%.

We also measure how varying model accuracy impacts the effect of model explanations, comparing four model accuracy conditions: 0\% (always incorrect), 50\%, 75\%, and 100\% (always correct)\footnote{We simulate desired model accuracy by post-hoc sub-sampling annotation instances where the model is correct/incorrect with corresponding frequency.}. When the model is always wrong, explanations reduce the human performance below both baselines (38\% in the EO condition, 26\% in ME). At 50\% model accuracy, human performance is higher than the ``random'' condition, but lower than the baseline. At 75\% , the ME condition outperforms the baseline (67\%). Finally, at 100\% model performance both model conditions improve the accuracy of the human annotation, with ME condition reaching 84\%. Our sample size of 540 exceeds the necessary 70 to holds a statistical power for between-subject studies \cite{bojko} .

Results from this experiment demonstrate that case-based explanations can improve human performance compared to a random baseline. However, the utility of the explanations is a function of the model accuracy. 


\subsection{Model Understanding} 
The second user task 
investigates model understanding by simulatability \cite{hase2019interpretable}: can the participant predict the model decision given the most important evidence consulted by the model? Specifically, we show five training examples to the user, either \textbf{Random Examples (RE)} or {\bf \tool{} Examples (PE)} (i.e., the same training examples used in the EO condition above). We ask participants to predict the model's decision using the same 5-point likert-scale as earlier. 

%



\paragraph{Results}
Per Figure \ref{fig:mu}, 
\tool{}'s explanations help the users predict the model behavior better than random examples: 50\% correct user assessment for PE vs 43.3\% for RE. In 23.3\% of the RE examples users are unable to make a prediction vs.\ 8\% for the PE. Random
guessing would be 40\% accurate on a five-way rating task with 2 positive, 1 neutral,
and 2 negative options (\S\ref{recogProp}). 

In Figure \ref{fig:mu-breakdown} we can see that the users are better at assessing the model prediction when the model is right (57\%) vs when the model is wrong (43\%). Additionally, we see that less users report inability to identify mode prediction when the model is correct (3.33\%) vs. when the model is not (13.3\%).


\section{Conclusion}

\tool{} is a novel approach to faithfully explain classification decisions by directly connecting model decisions with training examples via learned prototypes. \tool{} builds upon the state of the art in NLP. It integrates an underlying transformer encoder with prototype classification networks, and uses a novel, interleaving training algorithm for prototype learning. On the challenging propaganda detection task, \tool{} performed on-par in classification as its underlying encoder (BART-large), and exceeded BERT-large, with the added benefit of providing faithful model explanations via prototypes. 
Our pilot human evaluation study shows that additional input provided by \tool{} contains relevant information for the task and can improve the annotation performance, provided sufficient model accuracy. We further demonstrate that explanations help non-expert users better understand and simulate model predictions. 

\section*{Ethical Statement}

For annotation, we source participants from Amazon Mechanical Turk only within the United States, paying \$10/hour based on average task time. We did not reject any work but exclude data from participants who failed an attention check. 
\section*{Acknowledgements}

We thank the reviewers for their valuable feedback,  the online workers who participated in our study and provided annotations, and the Texas Advanced Computing Center (TACC) at UT Austin for its computational resources. This research was supported in part by NSF grants IIS-1850153 and IIS-2107524, as well as by Wipro, the Knight Foundation, the Micron Foundation, and by Good Systems\footnote{\url{http://goodsystems.utexas.edu/}}, a UT Austin Grand Challenge to develop responsible AI technologies. The statements made herein are solely the opinions of the authors and do not reflect the views of the sponsoring agencies.

\bibliography{references}

\begin{thebibliography}{52}
\expandafter\ifx\csname natexlab\endcsname\relax\def\natexlab#1{#1}\fi

\bibitem[{Agirre et~al.(2013)Agirre, Cer, Diab, Gonzalez-Agirre, and
  Guo}]{agirre-etal-2013-sem}
Eneko Agirre, Daniel Cer, Mona Diab, Aitor Gonzalez-Agirre, and Weiwei Guo.
  2013.
\newblock \href {https://aclanthology.org/S13-1004} {*{SEM} 2013 shared task:
  Semantic textual similarity}.
\newblock In \emph{Second Joint Conference on Lexical and Computational
  Semantics (*{SEM}), Volume 1: Proceedings of the Main Conference and the
  Shared Task: Semantic Textual Similarity}, pages 32--43, Atlanta, Georgia,
  USA. Association for Computational Linguistics.

\bibitem[{Amershi et~al.(2019)Amershi, Weld, Vorvoreanu, Fourney, Nushi,
  Collisson, Suh, Iqbal, Bennett, Inkpen et~al.}]{amershi2019guidelines}
Saleema Amershi, Dan Weld, Mihaela Vorvoreanu, Adam Fourney, Besmira Nushi,
  Penny Collisson, Jina Suh, Shamsi Iqbal, Paul~N Bennett, Kori Inkpen, et~al.
  2019.
\newblock \href {https://dl.acm.org/doi/10.1145/3290605.3300233} {Guidelines
  for human-ai interaction}.
\newblock In \emph{Proceedings of the 2019 chi conference on human factors in
  computing systems}, pages 1--13.

\bibitem[{Atanasova et~al.(2020)Atanasova, Simonsen, Lioma, and
  Augenstein}]{Atanasova2020GeneratingFC}
Pepa Atanasova, Jakob~Grue Simonsen, Christina Lioma, and Isabelle Augenstein.
  2020.
\newblock \href {https://doi.org/10.18653/v1/2020.acl-main.656} {Generating
  fact checking explanations}.
\newblock In \emph{Proceedings of the 58th Annual Meeting of the Association
  for Computational Linguistics}, pages 7352--7364, Online. Association for
  Computational Linguistics.

\bibitem[{Bansal et~al.(2021)Bansal, Wu, Zhou, Fok, Nushi, Kamar, Ribeiro, and
  Weld}]{bansal2021does}
Gagan Bansal, Tongshuang Wu, Joyce Zhou, Raymond Fok, Besmira Nushi, Ece Kamar,
  Marco~Tulio Ribeiro, and Daniel Weld. 2021.
\newblock \href {https://dl.acm.org/doi/10.1145/3411764.3445717} {Does the
  whole exceed its parts? the effect of ai explanations on complementary team
  performance}.
\newblock In \emph{Proceedings of the 2021 CHI Conference on Human Factors in
  Computing Systems}, pages 1--16.

\bibitem[{Bastings et~al.(2019)Bastings, Aziz, and
  Titov}]{bastings2019interpretable}
Jasmijn Bastings, Wilker Aziz, and Ivan Titov. 2019.
\newblock \href {https://doi.org/10.18653/v1/P19-1284} {Interpretable neural
  predictions with differentiable binary variables}.
\newblock In \emph{Proceedings of the 57th Annual Meeting of the Association
  for Computational Linguistics}, pages 2963--2977, Florence, Italy.
  Association for Computational Linguistics.

\bibitem[{Berg-Kirkpatrick et~al.(2012)Berg-Kirkpatrick, Burkett, and
  Klein}]{berg2012empirical}
Taylor Berg-Kirkpatrick, David Burkett, and Dan Klein. 2012.
\newblock \href {https://aclanthology.org/D12-1091} {An empirical investigation
  of statistical significance in {NLP}}.
\newblock In \emph{Proceedings of the 2012 Joint Conference on Empirical
  Methods in Natural Language Processing and Computational Natural Language
  Learning}, pages 995--1005, Jeju Island, Korea. Association for Computational
  Linguistics.

\bibitem[{Bien and Tibshirani(2011)}]{bien2011prototype}
Jacob Bien and Robert Tibshirani. 2011.
\newblock \href {"https://doi.org/10.1214/11-AOAS495"} {Prototype selection for
  interpretable classification}.
\newblock \emph{The Annals of Applied Statistics}, pages 2403--2424.

\bibitem[{Bojko(2013)}]{bojko}
Agnieszka Bojko. 2013.
\newblock \emph{Eye Tracking the User Experience: A Practical Guide to
  Research}.

\bibitem[{Chen et~al.(2019)Chen, Li, Tao, Barnett, Rudin, and
  Su}]{chen2019looks}
Chaofan Chen, Oscar Li, Daniel Tao, Alina Barnett, Cynthia Rudin, and
  Jonathan~K Su. 2019.
\newblock \href {https://arxiv.org/pdf/1806.10574.pdf} {This looks like that:
  Deep learning for interpretable image recognition}.
\newblock \emph{Advances in Neural Information Processing Systems},
  32:8930--8941.

\bibitem[{Da~San~Martino et~al.(2021)Da~San~Martino, Cresci,
  Barr{\'o}n-Cede{\~n}o, Yu, Di~Pietro, and Nakov}]{martino2020survey}
Giovanni Da~San~Martino, Stefano Cresci, Alberto Barr{\'o}n-Cede{\~n}o,
  Seunghak Yu, Roberto Di~Pietro, and Preslav Nakov. 2021.
\newblock \href {https://dl.acm.org/doi/abs/10.5555/3491440.3492112} {A survey
  on computational propaganda detection}.
\newblock In \emph{Proceedings of the Twenty-Ninth International Conference on
  International Joint Conferences on Artificial Intelligence}, pages
  4826--4832.

\bibitem[{Da~San~Martino et~al.(2019)Da~San~Martino, Yu, Barr{\'o}n-Cede{\~n}o,
  Petrov, and Nakov}]{da2019fine}
Giovanni Da~San~Martino, Seunghak Yu, Alberto Barr{\'o}n-Cede{\~n}o, Rostislav
  Petrov, and Preslav Nakov. 2019.
\newblock \href {https://doi.org/10.18653/v1/D19-1565} {Fine-grained analysis
  of propaganda in news article}.
\newblock In \emph{Proceedings of the 2019 Conference on Empirical Methods in
  Natural Language Processing and the 9th International Joint Conference on
  Natural Language Processing (EMNLP-IJCNLP)}, pages 5636--5646, Hong Kong,
  China. Association for Computational Linguistics.

\bibitem[{Danilevsky et~al.(2020)Danilevsky, Qian, Aharonov, Katsis, Kawas, and
  Sen}]{danilevsky2020survey}
Marina Danilevsky, Kun Qian, Ranit Aharonov, Yannis Katsis, Ban Kawas, and
  Prithviraj Sen. 2020.
\newblock \href {https://aclanthology.org/2020.aacl-main.46} {A survey of the
  state of explainable {AI} for natural language processing}.
\newblock In \emph{Proceedings of the 1st Conference of the Asia-Pacific
  Chapter of the Association for Computational Linguistics and the 10th
  International Joint Conference on Natural Language Processing}, pages
  447--459, Suzhou, China. Association for Computational Linguistics.

\bibitem[{Demartini et~al.(2020)Demartini, Mizzaro, and
  Spina}]{demartini2020human}
Gianluca Demartini, Stefano Mizzaro, and Damiano Spina. 2020.
\newblock \href
  {https://www.damianospina.com/publication/demartini-2020-human/demartini-2020-human.pdf}
  {Human-in-the-loop artificial intelligence for fighting online
  misinformation: Challenges and opportunities}.
\newblock \emph{The Bulletin of the Technical Committee on Data Engineering},
  43(3).

\bibitem[{Devlin et~al.(2019)Devlin, Chang, Lee, and
  Toutanova}]{devlin2019bert}
Jacob Devlin, Ming-Wei Chang, Kenton Lee, and Kristina Toutanova. 2019.
\newblock \href {https://doi.org/10.18653/v1/N19-1423} {{BERT}: Pre-training of
  deep bidirectional transformers for language understanding}.
\newblock In \emph{Proceedings of the 2019 Conference of the North {A}merican
  Chapter of the Association for Computational Linguistics: Human Language
  Technologies, Volume 1 (Long and Short Papers)}, pages 4171--4186,
  Minneapolis, Minnesota. Association for Computational Linguistics.

\bibitem[{Doshi-Velez and Kim(2017)}]{doshi2017towards}
Finale Doshi-Velez and Been Kim. 2017.
\newblock Towards a rigorous science of interpretable machine learning.
\newblock \emph{arXiv preprint arXiv:1702.08608}.

\bibitem[{Elbayad et~al.(2018)Elbayad, Besacier, and
  Verbeek}]{elbayad2018pervasive}
Maha Elbayad, Laurent Besacier, and Jakob Verbeek. 2018.
\newblock \href {https://doi.org/10.18653/v1/K18-1010} {Pervasive attention:
  2{D} convolutional neural networks for sequence-to-sequence prediction}.
\newblock In \emph{Proceedings of the 22nd Conference on Computational Natural
  Language Learning}, pages 97--107, Brussels, Belgium. Association for
  Computational Linguistics.

\bibitem[{Fomin et~al.(2020)Fomin, Anmol, Desroziers, Kriss, and
  Tejani}]{pytorch-ignite}
V.~Fomin, J.~Anmol, S.~Desroziers, J.~Kriss, and A.~Tejani. 2020.
\newblock High-level library to help with training neural networks in pytorch.
\newblock \url{https://github.com/pytorch/ignite}.

\bibitem[{Gad-Elrab et~al.(2019)Gad-Elrab, Stepanova, Urbani, and
  Weikum}]{GadElrab2019ExFaKTAF}
Mohamed~H. Gad-Elrab, Daria Stepanova, Jacopo Urbani, and Gerhard Weikum. 2019.
\newblock \href {https://doi.org/10.1145/3289600.3290996} {Exfakt: A framework
  for explaining facts over knowledge graphs and text}.
\newblock In \emph{Proceedings of the Twelfth ACM International Conference on
  Web Search and Data Mining}, WSDM '19, page 87–95, New York, NY, USA.
  Association for Computing Machinery.

\bibitem[{Glockner et~al.(2020)Glockner, Habernal, and
  Gurevych}]{glockner2020you}
Max Glockner, Ivan Habernal, and Iryna Gurevych. 2020.
\newblock \href {https://doi.org/10.18653/v1/2020.findings-emnlp.97} {Why do
  you think that? exploring faithful sentence-level rationales without
  supervision}.
\newblock In \emph{Findings of the Association for Computational Linguistics:
  EMNLP 2020}, pages 1080--1095, Online. Association for Computational
  Linguistics.

\bibitem[{Gold et~al.(2019)Gold, Kovatchev, and
  Zesch}]{gold-etal-2019-annotating}
Darina Gold, Venelin Kovatchev, and Torsten Zesch. 2019.
\newblock \href {https://doi.org/10.18653/v1/W19-4004} {Annotating and
  analyzing the interactions between meaning relations}.
\newblock In \emph{Proceedings of the 13th Linguistic Annotation Workshop},
  pages 26--36, Florence, Italy. Association for Computational Linguistics.

\bibitem[{Guu et~al.(2018)Guu, Hashimoto, Oren, and Liang}]{guu2018generating}
Kelvin Guu, Tatsunori~B. Hashimoto, Yonatan Oren, and Percy Liang. 2018.
\newblock \href {https://doi.org/10.1162/tacl_a_00030} {Generating sentences by
  editing prototypes}.
\newblock \emph{Transactions of the Association for Computational Linguistics},
  6:437--450.

\bibitem[{Hase and Bansal(2020)}]{hase2020evaluating}
Peter Hase and Mohit Bansal. 2020.
\newblock \href {https://doi.org/10.18653/v1/2020.acl-main.491} {Evaluating
  explainable {AI}: Which algorithmic explanations help users predict model
  behavior?}
\newblock In \emph{Proceedings of the 58th Annual Meeting of the Association
  for Computational Linguistics}, pages 5540--5552, Online. Association for
  Computational Linguistics.

\bibitem[{Hase et~al.(2019)Hase, Chen, Li, and Rudin}]{hase2019interpretable}
Peter Hase, Chaofan Chen, Oscar Li, and Cynthia Rudin. 2019.
\newblock \href {https://arxiv.org/abs/1906.10651} {Interpretable image
  recognition with hierarchical prototypes}.
\newblock In \emph{Proceedings of the AAAI Conference on Human Computation and
  Crowdsourcing}, volume~7, pages 32--40.

\bibitem[{Jacovi and Goldberg(2020)}]{jacovi2020towards}
Alon Jacovi and Yoav Goldberg. 2020.
\newblock \href {https://doi.org/10.18653/v1/2020.acl-main.386} {Towards
  faithfully interpretable {NLP} systems: How should we define and evaluate
  faithfulness?}
\newblock In \emph{Proceedings of the 58th Annual Meeting of the Association
  for Computational Linguistics}, pages 4198--4205, Online. Association for
  Computational Linguistics.

\bibitem[{Jain and Wallace(2019)}]{jain2019attention}
Sarthak Jain and Byron~C. Wallace. 2019.
\newblock \href {https://doi.org/10.18653/v1/N19-1357} {{A}ttention is not
  {E}xplanation}.
\newblock In \emph{Proceedings of the 2019 Conference of the North {A}merican
  Chapter of the Association for Computational Linguistics: Human Language
  Technologies, Volume 1 (Long and Short Papers)}, pages 3543--3556,
  Minneapolis, Minnesota. Association for Computational Linguistics.

\bibitem[{Jain et~al.(2020)Jain, Wiegreffe, Pinter, and
  Wallace}]{jain2020learning}
Sarthak Jain, Sarah Wiegreffe, Yuval Pinter, and Byron~C. Wallace. 2020.
\newblock \href {https://doi.org/10.18653/v1/2020.acl-main.409} {{L}earning to
  faithfully rationalize by construction}.
\newblock In \emph{Proceedings of the 58th Annual Meeting of the Association
  for Computational Linguistics}, pages 4459--4473, Online. Association for
  Computational Linguistics.

\bibitem[{Kim et~al.(2014)Kim, Rudin, and Shah}]{kim2014bayesian}
Been Kim, Cynthia Rudin, and Julie~A Shah. 2014.
\newblock \href
  {https://proceedings.neurips.cc/paper/2014/file/390e982518a50e280d8e2b535462ec1f-Paper.pdf}
  {The bayesian case model: A generative approach for case-based reasoning and
  prototype classification}.
\newblock In \emph{Advances in Neural Information Processing Systems},
  volume~27. Curran Associates, Inc.

\bibitem[{Kolodner(1992)}]{kolodner1992introduction}
Janet~L Kolodner. 1992.
\newblock An introduction to case-based reasoning.
\newblock \emph{Artificial intelligence review}, 6(1):3--34.

\bibitem[{Kotonya and Toni(2020{\natexlab{a}})}]{kotonya2020explainable}
Neema Kotonya and Francesca Toni. 2020{\natexlab{a}}.
\newblock \href {https://doi.org/10.18653/v1/2020.coling-main.474} {Explainable
  automated fact-checking: A survey}.
\newblock In \emph{Proceedings of the 28th International Conference on
  Computational Linguistics}, pages 5430--5443, Barcelona, Spain (Online).
  International Committee on Computational Linguistics.

\bibitem[{Kotonya and Toni(2020{\natexlab{b}})}]{Kotonya2020ExplainableAF}
Neema Kotonya and Francesca Toni. 2020{\natexlab{b}}.
\newblock \href {https://doi.org/10.18653/v1/2020.emnlp-main.623} {Explainable
  automated fact-checking for public health claims}.
\newblock In \emph{Proceedings of the 2020 Conference on Empirical Methods in
  Natural Language Processing (EMNLP)}, pages 7740--7754, Online. Association
  for Computational Linguistics.

\bibitem[{Kutlu et~al.(2020)Kutlu, McDonnell, Elsayed, and
  Lease}]{Kutlu20-jair}
Mucahid Kutlu, Tyler McDonnell, Tamer Elsayed, and Matthew Lease. 2020.
\newblock \href {https://www.jair.org/index.php/jair/article/view/12012}
  {{Annotator Rationales for Labeling Tasks in Crowdsourcing}}.
\newblock \emph{Journal of Artificial Intelligence Research (JAIR)},
  69:143--189.

\bibitem[{Lei et~al.(2016)Lei, Barzilay, and Jaakkola}]{lei2016rationalizing}
Tao Lei, Regina Barzilay, and Tommi Jaakkola. 2016.
\newblock \href {https://doi.org/10.18653/v1/D16-1011} {Rationalizing neural
  predictions}.
\newblock In \emph{Proceedings of the 2016 Conference on Empirical Methods in
  Natural Language Processing}, pages 107--117, Austin, Texas. Association for
  Computational Linguistics.

\bibitem[{Lewis et~al.(2020)Lewis, Liu, Goyal, Ghazvininejad, Mohamed, Levy,
  Stoyanov, and Zettlemoyer}]{lewis2020bart}
Mike Lewis, Yinhan Liu, Naman Goyal, Marjan Ghazvininejad, Abdelrahman Mohamed,
  Omer Levy, Veselin Stoyanov, and Luke Zettlemoyer. 2020.
\newblock \href {https://doi.org/10.18653/v1/2020.acl-main.703} {{BART}:
  Denoising sequence-to-sequence pre-training for natural language generation,
  translation, and comprehension}.
\newblock In \emph{Proceedings of the 58th Annual Meeting of the Association
  for Computational Linguistics}, pages 7871--7880, Online. Association for
  Computational Linguistics.

\bibitem[{Li et~al.(2018)Li, Liu, Chen, and Rudin}]{li2018deep}
Oscar Li, Hao Liu, Chaofan Chen, and Cynthia Rudin. 2018.
\newblock \href {https://dl.acm.org/doi/abs/10.5555/3504035.3504467} {Deep
  learning for case-based reasoning through prototypes: A neural network that
  explains its predictions}.
\newblock In \emph{Proceedings of the Thirty-Second AAAI Conference on
  Artificial Intelligence and Thirtieth Innovative Applications of Artificial
  Intelligence Conference and Eighth AAAI Symposium on Educational Advances in
  Artificial Intelligence}. AAAI Press.

\bibitem[{Liao et~al.(2020)Liao, Gruen, and Miller}]{liao2020questioning}
Q~Vera Liao, Daniel Gruen, and Sarah Miller. 2020.
\newblock \href {https://dl.acm.org/doi/10.1145/3313831.3376590} {Questioning
  the {AI}: informing design practices for explainable ai user experiences}.
\newblock In \emph{Proceedings of the 2020 CHI Conference on Human Factors in
  Computing Systems}, pages 1--15.

\bibitem[{Loshchilov and Hutter(2019)}]{loshchilov2018decoupled}
Ilya Loshchilov and Frank Hutter. 2019.
\newblock \href {https://openreview.net/forum?id=Bkg6RiCqY7} {Decoupled weight
  decay regularization}.
\newblock In \emph{International Conference on Learning Representations}.

\bibitem[{Nakov et~al.(2021)Nakov, Corney, Hasanain, Alam, Elsayed,
  Barr'on-Cedeno, Papotti, Shaar, and Martino}]{Nakov2021AutomatedFF}
Preslav Nakov, D.~Corney, Maram Hasanain, Firoj Alam, Tamer Elsayed,
  A.~Barr'on-Cedeno, Paolo Papotti, Shaden Shaar, and Giovanni Da~San Martino.
  2021.
\newblock Automated fact-checking for assisting human fact-checkers.
\newblock In \emph{IJCAI}.

\bibitem[{Nguyen et~al.(2018)Nguyen, Kharosekar, Krishnan, Krishnan, Tate,
  Wallace, and Lease}]{nguyen2018believe}
An~T Nguyen, Aditya Kharosekar, Saumyaa Krishnan, Siddhesh Krishnan, Elizabeth
  Tate, Byron~C Wallace, and Matthew Lease. 2018.
\newblock Believe it or not: Designing a human-ai partnership for
  mixed-initiative fact-checking.
\newblock In \emph{Proceedings of the 31st Annual ACM Symposium on User
  Interface Software and Technology}, pages 189--199.

\bibitem[{Popat et~al.(2018)Popat, Mukherjee, Yates, and
  Weikum}]{Popat2018DeClarEDF}
Kashyap Popat, Subhabrata Mukherjee, Andrew Yates, and Gerhard Weikum. 2018.
\newblock \href {https://doi.org/10.18653/v1/D18-1003} {{D}e{C}lar{E}:
  Debunking fake news and false claims using evidence-aware deep learning}.
\newblock In \emph{Proceedings of the 2018 Conference on Empirical Methods in
  Natural Language Processing}, pages 22--32, Brussels, Belgium. Association
  for Computational Linguistics.

\bibitem[{Pruthi et~al.(2020)Pruthi, Gupta, Dhingra, Neubig, and
  Lipton}]{pruthi2020learning}
Danish Pruthi, Mansi Gupta, Bhuwan Dhingra, Graham Neubig, and Zachary~C.
  Lipton. 2020.
\newblock \href {https://doi.org/10.18653/v1/2020.acl-main.432} {Learning to
  deceive with attention-based explanations}.
\newblock In \emph{Proceedings of the 58th Annual Meeting of the Association
  for Computational Linguistics}, pages 4782--4793, Online. Association for
  Computational Linguistics.

\bibitem[{Rajagopal et~al.(2021)Rajagopal, Balachandran, Hovy, and
  Tsvetkov}]{rajagopal-etal-2021-selfexplain}
Dheeraj Rajagopal, Vidhisha Balachandran, Eduard~H Hovy, and Yulia Tsvetkov.
  2021.
\newblock \href {https://doi.org/10.18653/v1/2021.emnlp-main.64}
  {{SELFEXPLAIN}: A self-explaining architecture for neural text classifiers}.
\newblock In \emph{Proceedings of the 2021 Conference on Empirical Methods in
  Natural Language Processing}, pages 836--850, Online and Punta Cana,
  Dominican Republic. Association for Computational Linguistics.

\bibitem[{Ribeiro et~al.(2016)Ribeiro, Singh, and Guestrin}]{ribeiro2016should}
Marco~Tulio Ribeiro, Sameer Singh, and Carlos Guestrin. 2016.
\newblock " why should i trust you?" explaining the predictions of any
  classifier.
\newblock In \emph{Proceedings of the 22nd ACM SIGKDD international conference
  on knowledge discovery and data mining}, pages 1135--1144.

\bibitem[{Ribeiro et~al.(2020)Ribeiro, Wu, Guestrin, and
  Singh}]{ribeiro2020beyond}
Marco~Tulio Ribeiro, Tongshuang Wu, Carlos Guestrin, and Sameer Singh. 2020.
\newblock \href {https://doi.org/10.18653/v1/2020.acl-main.442} {Beyond
  accuracy: Behavioral testing of {NLP} models with {C}heck{L}ist}.
\newblock In \emph{Proceedings of the 58th Annual Meeting of the Association
  for Computational Linguistics}, pages 4902--4912, Online. Association for
  Computational Linguistics.

\bibitem[{Serrano and Smith(2019)}]{serrano2019attention}
Sofia Serrano and Noah~A. Smith. 2019.
\newblock \href {https://doi.org/10.18653/v1/P19-1282} {Is attention
  interpretable?}
\newblock In \emph{Proceedings of the 57th Annual Meeting of the Association
  for Computational Linguistics}, pages 2931--2951, Florence, Italy.
  Association for Computational Linguistics.

\bibitem[{Shu et~al.(2019)Shu, Cui, Wang, Lee, and Liu}]{shu2019defend}
Kai Shu, Limeng Cui, Suhang Wang, Dongwon Lee, and Huan Liu. 2019.
\newblock defend: Explainable fake news detection.
\newblock In \emph{Proceedings of the 25th ACM SIGKDD International Conference
  on Knowledge Discovery \& Data Mining}, pages 395--405.

\bibitem[{Sundararajan et~al.(2017)Sundararajan, Taly, and
  Yan}]{sundararajan2017axiomatic}
Mukund Sundararajan, Ankur Taly, and Qiqi Yan. 2017.
\newblock Axiomatic attribution for deep networks.
\newblock In \emph{International Conference on Machine Learning}, pages
  3319--3328. PMLR.

\bibitem[{Ulyanov et~al.(2016)Ulyanov, Vedaldi, and
  Lempitsky}]{ulyanov2016instance}
Dmitry Ulyanov, Andrea Vedaldi, and Victor Lempitsky. 2016.
\newblock Instance normalization: The missing ingredient for fast stylization.
\newblock \emph{arXiv preprint arXiv:1607.08022}.

\bibitem[{Wang et~al.(2019)Wang, Yang, Abdul, and Lim}]{wang2019designing}
Danding Wang, Qian Yang, Ashraf Abdul, and Brian~Y Lim. 2019.
\newblock Designing theory-driven user-centric explainable {AI}.
\newblock In \emph{Proceedings of the 2019 CHI conference on human factors in
  computing systems}, pages 1--15.

\bibitem[{Wang et~al.(2021)Wang, Choi, Xu, and Yang}]{wang2021putting}
Zijie~J. Wang, Dongjin Choi, Shenyu Xu, and Diyi Yang. 2021.
\newblock \href {https://aclanthology.org/2021.hcinlp-1.8} {Putting humans in
  the natural language processing loop: A survey}.
\newblock In \emph{Proceedings of the First Workshop on Bridging
  Human{--}Computer Interaction and Natural Language Processing}, pages 47--52,
  Online. Association for Computational Linguistics.

\bibitem[{Wickramasinghe et~al.(2020)Wickramasinghe, Marino, Grandio, and
  Manic}]{wickramasinghe2020trustworthy}
Chathurika~S Wickramasinghe, Daniel~L Marino, Javier Grandio, and Milos Manic.
  2020.
\newblock \href {https://ieeexplore.ieee.org/document/9142644} {Trustworthy
  {AI} development guidelines for human system interaction}.
\newblock In \emph{2020 13th International Conference on Human System
  Interaction (HSI)}, pages 130--136. IEEE.

\bibitem[{Wiegreffe and Pinter(2019)}]{wiegreffe2019attention}
Sarah Wiegreffe and Yuval Pinter. 2019.
\newblock \href {https://doi.org/10.18653/v1/D19-1002} {Attention is not not
  explanation}.
\newblock In \emph{Proceedings of the 2019 Conference on Empirical Methods in
  Natural Language Processing and the 9th International Joint Conference on
  Natural Language Processing (EMNLP-IJCNLP)}, pages 11--20, Hong Kong, China.
  Association for Computational Linguistics.

\bibitem[{Xu et~al.(2015)Xu, Ba, Kiros, Cho, Courville, Salakhudinov, Zemel,
  and Bengio}]{xu2015show}
Kelvin Xu, Jimmy Ba, Ryan Kiros, Kyunghyun Cho, Aaron Courville, Ruslan
  Salakhudinov, Rich Zemel, and Yoshua Bengio. 2015.
\newblock \href {http://proceedings.mlr.press/v37/xuc15.pdf} {Show, attend and
  tell: Neural image caption generation with visual attention}.
\newblock In \emph{International conference on machine learning}, pages
  2048--2057. PMLR.

\end{thebibliography}
\bibliographystyle{acl_natbib}

\appendix

\section{Appendix} 

\subsection{Prototypes as Soft-Clustering} \label{subsec:math}

We provide more insights into prototypes by illustrating how the prototype layer relates  to soft-clustering . 

Let $t_{1:n}$ denote $n$ training examples, having binary labels $y_{1:n}$. Let $D(a,b)$ denote symmetric distance of any two training examples $a$ and $b$. Assume $m$ additional {\em prototypes} (i.e., points) $p_{1:m}$ are defined in the same space as the training examples. Then $D(a,b)$ can also be computed between any training example and prototype, or between any two prototypes. Let $d^j_i = D(p_j,t_i)$ denote the symmetric distance between $p_j$ and training example $t_i$.  Then any two training examples, $t_u$ and $t_v$, will have respective distances $d^j_u$ and $d^j_v$ to prototype $p_j$. 

Let $P_j = \pi^j_{1:n}$ denote a probability distribution for prototype $p_j$ over the training examples $t_{1:n}$. Specifically, induce $\pi^j_i$ for training example $t_i$ as a function of its distance $d^j_i$ from prototype $p_j$: $\pi^j_i = z_j/d^j_i$, where $z_j$ is a normalization constant. Then the relative probabilities for two training examples $t_v$ and $t_u$ = $\pi^j_v / \pi^j_u = \frac{z_j/d^j_v}{z_j/d^j_u} = d^j_u / d^j_v$. By total probability, $1 = \sum_i \pi^j_i = \sum_i z_j/d^j_i = z_j \sum_i 1/d^j_i$, so $z_j = \frac{1}{\sum_i 1/d^j_i}$. Based on this, we can say that each prototype effectively denotes a {\em soft-clustering} over the set of training examples. 

Further, the ratio of distances ($d^j_u$ / $d^j_v$) between training examples $t_u$ and $t_v$ and prototype $p_j$, is the reciprocal of their probabilities: $\pi^j_v / \pi^j_u$. In other words, if a training example $t_u$ is twice as far away from prototype $p_j$ as another training example $t_v$ (i.e., $d^j_u$ / $d^j_v = 2$), then $t_v$ will be twice as probable as $t_u$ in probability distribution $P_j$ (i.e., $\pi^j_v / \pi^j_u = 2$).

\paragraph{Inference.} The inference calculation shown here uses only the prototype layer.
$P_j(y=1) = \psi_j = \sum_{1:n} \pi^j_i y_i$ denote the relative frequency estimated probability of prototype $p_j$ having true class label $y=1$. Let $x$ denote a test example (defined in the same vector space as training examples and prototypes). Then $D(x,p_j) = d^j_x$ defines the symmetric distance between $x$ and prototype $p_j$. Let $\Theta_x = \theta^x_{1:m}$ denote a probability distribution for test example $x$ over the prototypes $p_{1:m}$. As with training examples and prototypes above, induce this probability distribution based on relative distances between $x$ and each prototype $p_j$. Then similar to before, if $d^j_x / d^k_x = 2$, meaning prototype $p_j$ is twice as far from $x$ as prototype $p_k$, then we have $\theta^x_k / \theta^x_j = 2$ meaning $p_k$ will be twice as probable as $p_j$ in probability distribution $\Theta_x$. Class label $y=1$ for test example $x$ is predicted by probability $\Theta_x(y=1) = \sum_{1:m} \theta^x_j \psi_j = \sum_{j\in1:m} \theta^x_j \sum_{i\in1:n}  \pi^j_i y_i$.

\end{document}